\documentclass{article}
\usepackage{arxiv}

\usepackage{pgfplots}
\pgfplotsset{width=10cm, compat=newest}

\newcommand{\dataset}{{PERLEX}}

\title{\dataset: A Bilingual Persian-English Gold Dataset for Relation Extraction}

\author{
    Majid Asgari-Bidhendi\\
    School of Computer Engineering\\
    Iran University of Science and Technology\\
    Tehran, Iran\\
    \texttt{majid.asgari@gmail.com}\\
    \And
    Mehrdad Nasser\\
    School of Computer Engineering\\
    Iran University of Science and Technology\\
    Tehran, Iran\\
    \texttt{Mehrdad\_naser\_73@yahoo.com}\\
    \And
    Behrooz Janfada\\
    School of Computer Engineering\\
    Iran University of Science and Technology\\
    Tehran, Iran\\
    \texttt{behrooz.janfada@gmail.com}\\
    \And
    Behrouz Minaei-Bidgoli\\
    School of Computer Engineering\\
    Iran University of Science and Technology\\
    Tehran, Iran\\
    \texttt{b\_minaei@iust.ac.ir}\\
}

\begin{document}
\maketitle
\begin{abstract}
    Relation extraction is the task of extracting semantic relations between entities in a sentence. It is an essential part of some natural language processing tasks such as information extraction, knowledge extraction, and knowledge base population. The main motivations of this research stem from a lack of a dataset for relation extraction in the Persian language as well as the necessity of extracting knowledge from the growing big-data in the Persian language for different applications. In this paper, we present ``\dataset'' as the first Persian dataset for relation extraction, which is an expert-translated version of the ``Semeval-2010-Task-8'' dataset. Moreover, this paper addresses Persian relation extraction utilizing state-of-the-art language-agnostic algorithms. We employ six different models for relation extraction on the proposed bilingual dataset, including a non-neural model (as the baseline), three neural models, and two deep learning models fed by multilingual-BERT contextual word representations. The experiments result in the maximum f-score 77.66\% (provided by BERTEM-MTB method) as the state-of-the-art of relation extraction in the Persian language.
\end{abstract}

\section{Introduction}
Relation Extraction (RE) is the task of identifying semantic relations between text entities and is one of the most crucial tasks in Natural Language Processing (NLP). In RE, entities are string literals which are marked in the sentence. Furthermore, the goal in RE is to detect a limited number of pre-defined relationships from the text. 
Knowledge base population is one of the applications of RE. A knowledge base contains a set of entities and relationships between them. There are many knowledge bases available in English at the moment, such as Yago  \cite{suchanek2007yago}, Freebase \cite{bollacker2008freebase}, DBpedia \cite{auer2007dbpedia} and Wikidata \cite{vrandevcic2014wikidata}. However, the first knowledge base in the Persian language was developed recently \cite{asgari2019farsbase}, which was one of the motivations for this research.

At the outset, the first dataset for the Persian RE is introduced. Then, five language-agnostic methods for the RE task are employed, and the results of the method are compared with a baseline.

Although there are already standard RE datasets in English such as Semeval-2010-Task-8 (Multi-Way Classification of Semantic Relations Between Pairs of Nominals) \cite{hendrickx2009semeval}, TACRED \cite{zhang2017position}, and ACE 2005 \cite{walker2006ace}, there is no dataset available for the Persian language. Thus, in this paper, we present ``\dataset'', which is an expert-translated version of the Semeval-2010-Task-8 dataset.

The evaluation and comparison of the selected RE methods are carried out using \dataset\footnote{\dataset~dataset is available at http://farsbase.net/PERLEX.html} dataset. It is rational to adapt existing state-of-the-art language-independent RE methods with our target language, i.e. Persian by reimplementation. We use five neural RE models, including a model based on convolutional neural networks, two models based on recurrent neural networks and two BERT-based models. 

Before the introduction of BERT \cite{devlin2019bert}, the BLSTM-LET \cite{lee2019semantic} model was one of the best models presented for the RE task. The application of Bidirectional LSTM Networks with Entity-aware Attention using Latent Entity Typing (BLSTM-LET) method for the RE task was regarded as one of the state-of-the-art language-agnostic approaches. BLSTM-LET outperformed previous state-of-the-art RE methods without the use of customized linguistic features while it relies solely on word embedding \cite{mikolov2013efficient} features.

With the advent of BERT, many tasks of NLP have evolved. BERT \cite{devlin2019bert} is a contextual text representation model that was shown to achieve state-of-the-art results in 11 different NLP tasks. Unlike previous word representations where each word would have a fixed embedding, words have different embeddings with BERT in different contexts. At present, the BERTEM-MTB \cite{soares2019matching} model has shown that in both Semeval-2010-Task-8 and TACRED datasets, it is the state-of-the-art of the RE task.

The remainder of this paper is organized as follows. Section~\ref{relatedWorks} provides a summary of the related literature in RE task. In section~\ref{dataset}, we elaborate on the design of our proposed dataset, \dataset. Section~\ref{re-experiments}  presents the experimental results along with further analyses of the obtained results. Finally, in section~\ref{conclusion}, we conclude this paper and propose possible future lines of extension of this study.
\section{Related Works}
\label{relatedWorks}
In the following section, we first provide a brief review of the well-known RE datasets. Then, we divide the state-of-the-art RE algorithms into two different categories: deep-learning-based methods and non-deep-learning-based methods. The performance of these models on the \dataset~dataset is reported in the section~\ref{re-experiments}.
\subsection{Datasets}
RE datasets can be classified into two general groups: distantly-supervised datasets and hand-labelled datasets.
    
In the hand-labelled datasets, the label of each relation mention is determined by human experts. Thus, the creation of such datasets is time-consuming and expensive. Datasets like ACE \cite{walker2006ace}, Semeval-2010-Task-8 \cite{hendrickx2009semeval}, TACRED \cite{zhang2017position}, and FewRel \cite{han2018fewrel} belong to the hand-labeled category.

However, labels of relation mentions in the distantly-supervised datasets are determined by the corresponding relations of the mentioned pairs in a knowledge base. The most extensive use of distantly-supervised dataset is in the approach proposed by Mintz et al. \cite{mintz2009distant}. NYT-10 \cite{riedel2010modeling} where entities are aligned in the New York Times corpus to entities in Freebase. 

Distantly-supervised datasets have Some advantages over hand-labelled ones. For example, there is no need for human experts to do the time-consuming process of annotation on the distantly-supervised datasets. Furthermore, distantly-supervised datasets can utilize labels already used in knowledge bases, which makes such datasets ideal for knowledge-base-related tasks such as knowledge base population. The main disadvantage of these datasets is their noisy labels.
The are many approaches proposed to deal with the problem of noisy labels such as multiple-instance learning \cite{riedel2010modeling,hoffmann2011knowledge,surdeanu2012multi}, reinforcement learning \cite{yang2019exploiting,qin2018robust,feng2018reinforcement}, the use of knowledge base side information \cite{vashishth2018reside,wang2018label}, and attention mechanism \cite{lin2016neural,ye2019distant}. In the Persian language, FarsBase \cite{asgari2019farsbase} especially uses a distant-supervised method to extract triples for the knowledge base.
    
\subsection{Non-Deep-Learning-Based Methods}
Before the advent of deep learning models, NLP tasks relied on specific NLP tools such as dependency parsers and POS taggers for feature extraction. These models are not able to compete with deep learning models due to the costly nature of their handcrafted features and resources. Nevertheless, these features generally are extracted by NLP tools while some errors may cause by themselves. These methods employ classifiers, such as SVM and Maximum Entropy (MaxEnt). The-state-of-the-art method for RE was achieved by Rink and Harabagiu \cite{rink2010utd} on Semeval-2010-Task-8 dataset in 2010 using an SVM with several handcrafted features and resources including Lexical resources, Dependency, PropBank, FrameNet, Hypernym, NumLex-Plus, NGrams, and TextRunner \cite{yates2007textrunner}. Their model was the best non-deep-learning-based model, however later, it was outperformed by deep-learning-based models such as CNN and RNN methods.

LightRel \cite{renslow2018lightrel} is another non-deep-learning-based method, which is a fast and lightweight logistic regression classifier. In this method, a relation mention is represented as a sequence of tokens. The main idea of this method consists of transforming these sequences into vectors of fixed length such that each token (or word) is represented only by four features including the word itself, its shape (a small, fixed amount of character-based features), the word’s cluster-id extracted from external knowledge base, and the word’s embedding of fixed size. Then a logistic regression classification model is trained to predict classes using feature vectors.

\subsection{Deep-Learning-Based Methods}
In this section, we present and describe some of the essential features of deep-learning-based models used for RE. Each of these models was state-of-the-art at its time, but shortly afterwards, it was outperformed by the next model.
    
\textbf{Convolutional Neural Networks (CNNs)} were initially used in computer vision to extract features from images, but they have recently been applied to various NLP tasks. Zeng et al. \cite{zeng2014relation} used CNNs to extract features from sentences and classified their relations using these features. Their proposed model used a set of convolution and pooling layers followed by two fully-connected layers and a Softmax classifier to classify relations. Features of each word are the concatenation of word and position embeddings. Word embeddings are the vector representations of words in a d-dimensional space, i.e. a vector with the size of d representing a word in each dimension. Position embeddings are the relative position of each word in a sentence relevant to the two entities in the sentence. Intuitively, what the convolution layer does, is the encoding of every N consecutive word into a feature vector while N is the kernel length of the convolution layer and is a hyper-parameter. A max-pooling layer is then used to extract the most relevant features of a sentence. Next, these feature vectors are fed into two fully-connected layers, and a Softmax classifier is used to determine the relation.  
    
\textbf{Attention-based Bidirectional Long Short-Term Memory Network (Att-BLSTM)} is a RE model proposed by Zhou et al. \cite{zhou2016attention}, which is capable of surpassing many state-of-the-art models without relying on NLP tools or lexical resources for feature extraction. This model utilizes a Recurrent Neural Network (RNN) for classification. In a regular RNN, the output of each time step is dependent on the current input and the output of the previous time step. However, in some tasks (e.g., machine translation), the output of the current time step is also dependent on the outputs of the future time steps. In such cases, bidirectional RNNs can be utilized. In this model, the embedding of each word is given as an input to two LSTM cells, one of which for the forward pass and the other for the backward pass. The outputs of each pair are then concatenated to produce the output relevant to each word. Following this LSTM layer, an attention layer is employed, producing the network's output by constructing a weighted sum of all outputs relevant to each word. The attention layer's function, as the name suggests, denotes higher weights to words with higher importance, which results in distinguishing more important words from others. For instance, the word "the" is less useful than a word like "Caused" for determining the Cause-Effect relation in a sentence. These attention weights are learned in the process of training. Should be noted, the Att-BLSTM model is independent of lexical or syntactic features and relies solely on word embeddings. This model was capable of outperforming many state-of-the-art models that rely on features extracted by NLP tools on the SemEval- 2010-task 8 dataset.
    
\textbf{Bidirectional LSTM with Entity-aware Attention using Latent Entity Typing (BLSTM-LET)} proposed by Lee et al. \cite{lee2019semantic}, utilizes the self-attention introduced by Vaswani et al. \cite{vaswani2017attention}, and latent entity typing to produce better representations of words. Additionally, A bi-directional LSTM is used for classification as well. Word embeddings are used as the input of the model. Multi-head attention is then used to produce contextualized representations for the words. These representations are then concatenated to position embeddings and latent entity type embeddings and fed into an attention layer to obtain the final representation of the sentence. Latent entity types provide information about the entities. Intuitively, latent entity types are clusters to which an entity can belong. Latent entity type embeddings are learned during the training process through the contextualized representation of the entities \cite{lee2019semantic}. When dimension reduction is applied to entity type embeddings with visualization of entities in a 2-d plot, it can be shown that entity pairs such as ``pollution'' and ``virus'', or ``worker'' and ``chairman'' fall into the same cluster. This model outperformed any models that did not use NLP tools to extract features except BERT-based models.

BERT-based models have recently been applied in the field of RE and have been able to obtain the best results up to now and outperform previous methods.

In contrast to context-free models such as word2vec, Bidirectional Encoder Representations from Transformers (BERT) \cite{devlin2019bert} is an unsupervised context-dependent language representation model. BERT was shown to achieve state-of-the-art results in different NLP tasks such as a set of eight tasks named GLUE when it was introduced. As opposed to previous word representations where each word would have a single fixed embedding vector, BERT embedding word vectors are different in any context. RE was not among the tasks where BERT has experimented; however, BERT discovered its path through this field immediately. 

\textbf{Enriching Pre-trained Language Model with Entity Information (R-BERT)} is a recently proposed model by Wu and He \cite{wu2019enriching}, which used BERT for the task of RE and showed to be the best method on the Semeval-2010-Task-8 dataset. To encode a relation between two entities of a sentence using BERT, R-BERT adds the special token ``\$'' before and after the first entity and another special token ``\#'' before and after the second entity in a given sentence. R-BERT also adds another special token ``[CLS]'' to the beginning of each sentence. The final representation of each relation is calculated by concatenating three hidden state vectors including the hidden state vector corresponding to the [CLS] token, the averages of the hidden state vectors corresponding to the first, and second entity tokens. A fully connected layer followed by a softmax layer is then used to map the acquired relation representation to a relation.

\textbf{BERTEM with Matching the Blanks (BERTEM-MTB)} is the most recent and current state-of-the-art approach, which is a BERT-based model and is very similar to R-BERT. This method was proposed by Soares et al. \cite{soares2019matching}. Similar to R-BERT, BERTEM-MTB method added special tokens before and after the entities. Unlike R-BERT, the tokens before and after each entity are different in BERTEM-MTB method. Relation representation in this model is the concatenation of the hidden states of the special tokens before each entity. These relation representations are then used to classify each sentence's relation. This method also adds another training step in the architecture of fine-tuning relation representations in BERT by replacing the entities with ``[BLANK]'' special token in sentences which their entity pairs are similar.
\section{Construction of the \dataset~bilingual dataset}
\label{dataset}
Unlike English and other rich-resource languages, the Persian language has no proper assets available for RE. In the English language, Semeval-2010-Task-8 challenge is one of the most well-known datasets for RE which has been utilized in many studies. This dataset contains 10717 example sentences and their corresponding relation types from which 8000 is for training and 2717 for the test. In this challenge, each relation extraction algorithm is asked to identify one of the nine predefined relationships in this dataset for the pair of entities specified in each sentence. The nine predefined relations are ``Cause-Effect'', ``Component-Whole'', ``Content-Container'', ``Entity-Destination'', ``Entity-Origin'', ``Instrument-Agency'', ``Member-Collection'', ``Message-Topic'', ``Product-Producer, and the relation ``Other'' in case there is not a confirmed relationship between the two entities.
    
\dataset~is the parallel translation of all examples in Semeval-2010-Task-8 dataset. With this approach, the cost of sentence selection is eliminated. On the other hand, this dataset is constructed from an original and widely-utilized dataset. Therefore, it is possible to implicitly compare the results of implementing RE methods on this dataset with those of the English dataset. Table~\ref{corpus-tbl} illustrates the statistics related to the dataset.

\begin{table*}
    \caption{\dataset~dataset Statistics}
    \label{corpus-tbl}
	\begin{tabular}{@{\extracolsep{\fill}}lrrrrrr}
		\hline
		Partition & Examples & Words   &  Args  & Arg Words & Avg. Sentence Length & Matched Arg with EL \\
		\hline
		Train   & 8,000    & 160,266 & 16,000 & 20,697    & 20.03           & 68.76\%             \\
		
		Test    & 2,717    & 54,351  & 5,434  & 6,679     & 20.00           & 67.66\%             \\
		\hline
		Total   & 10,717   & 21,4617 & 21,434 & 27,376    & 20.02           & 68.48\%             \\
		\hline
	\end{tabular}
\end{table*}
\section{Experiments and Analyses of Relation Extraction}
\label{re-experiments}
In this section, we report experimental settings and classification results of six different models: Baseline, CNN, Att-BLSTM, BLSTM-LET, R-BERT, and BERTEM-MTB. 
\subsection{Experimental Setup}
In \dataset, we adapted the nine datasets similar to those in Semeval-2010-Task-8 dataset mentioned in section~\ref{relatedWorks}. Each class has two variations specifying the placement order of the subject and object in the sentence. For instance, Cause-Effect class possesses two variants: Cause-Effect (e1,e2) and Cause-Effect (e2,e1).
    
Generally, there are three ways to evaluate the classification results:
\begin{enumerate}
	\item Taking into account both variations of each class (18 classes in total).
	\item Using only one variation of each class (and considering directionality).
	\item Using only one variation of each class (and ignoring directionality).
\end{enumerate}

Moreover, there are two methods to measure f1-score, namely Micro-averaging and Macro-averaging. Additionally, the pairs of entities that do not fall into any of the main nine classes are labelled as ``Other'' in the dataset and do not participate in evaluations. We adopt the official evaluation method for Semeval-2010-Task-8 dataset, which is (9+1)-way classification with macro-averaging f1-score measurement while directionality is taken into account. This (9+1)-way means that we use the nine main classes plus ``Other'' in training and testing but ``Other'' is ignored when we calculate the f1-scores. In all non-BERT-based experiments, we use 300-dimensional word embeddings pre-trained by Poostchi et al. \cite{poostchi2016personer} and utilize 10\% of the training set as the development set. 

\subsection{Overall Results}
\begin{figure}[t]
\centerline{\vbox to 5pc{\hbox to 6pc{}}}
\begin{tikzpicture}
\begin{axis}[
    legend style={at={(0.5,1.15)},
    anchor=north,legend columns=0},
    xtick={0,1,2,3,4,5,6,7,8,9,10,11},
    xticklabels={
        {LRC},
        {LRC},
        {CNN},
        {CNN},
        {Att-BLSTM},
        {Att-BLSTM},
        {BLSTM-LET},
        {BLSTM-LET},
        {R-BERT},
        {R-BERT},
        {BERTEM-MTB},
        {BERTEM-MTB}},
    nodes near coords,
    axis lines*=left,
    ymin=0.0,
    ylabel = Official f1-score,
    x tick label style={rotate=270},
    ybar=2pt,
    bar shift=0pt,
    height=6cm,
    width=1\textwidth,
    ]
    \addplot+[ybar] plot coordinates {
        (1, 57.42)
        (3, 69.28)
        (5, 69.61)
        (7, 70.79)
        (9, 75.31)
        (11, 77.66)
        };
    \addplot+[ybar] plot coordinates {
        (0, 79.78)
        (2, 82.70)
        (4, 84.00)
        (6, 85.20)
        (8, 89.25)
        (10, 89.50)
        };
    \legend{Persian, English}
\end{axis}
\end{tikzpicture}
\caption{f1-score for all models}
\label{f1-score-for-all-models-fig}
\end{figure}
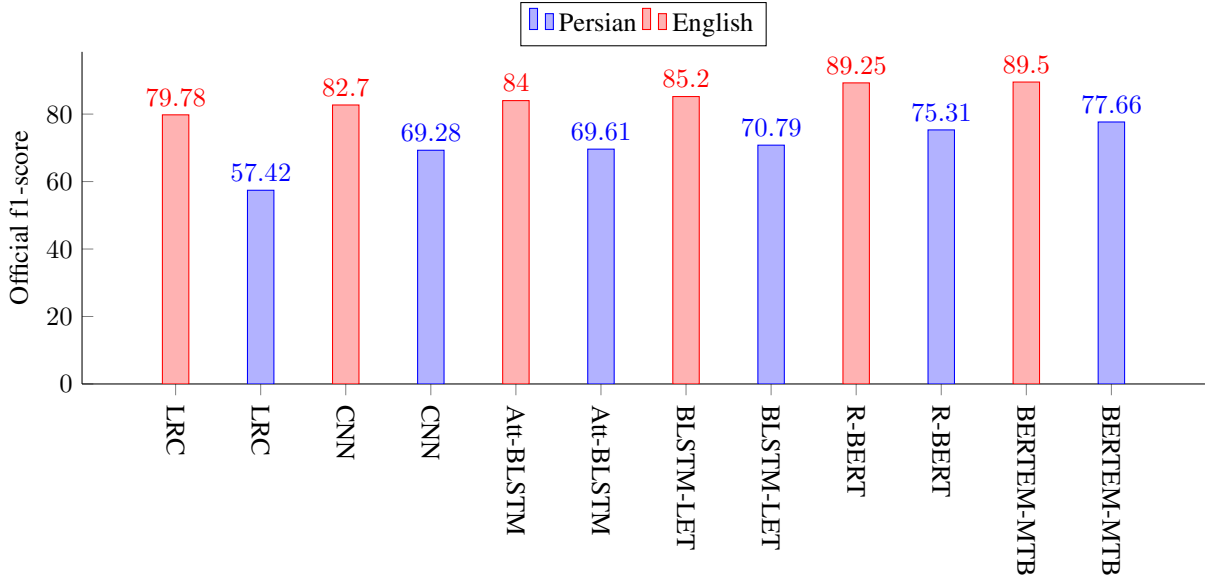

Figure~\ref{f1-score-for-all-models-fig} illustrates the official f1-scores for each model. As we expected, the results are lower for the Persian language in comparison with the English language. This drawback is due to many challenges in the processing of the Persian language as a free-word-order and a more ambiguous language. Likewise, in English, the performance of BERTEM-MTB overcomes that of all the other five methods in the Persian language. Moreover, BERTEM is the superior method in all nine classes in the Persian language.

\subsubsection{Baseline}
We use features to train a Logistic Regression Classifier (LRC) and  L2R\_LR solver as the baseline for the Persian language such as Word IDs (unique IDs for each word in the dataset), Part of Speech (POS), Tags for each pair of entities, Words between two entities, POS tags of the words between two entities, Dependency relations and their direction between two consecutive entities, POS word tags between two entities and The bag of words.

Based on the obtained results, the official f1-score for logistic regression on \dataset~dataset is 57.42\%. It should be noted that we report logistic regression baseline method of LightRel \cite{renslow2018lightrel} results for English.
    
\subsubsection{CNN}
For the CNN model, we use four different kernel lengths: 2, 3, 4, and 5. Then, we concatenate the outputs of these kernels. We set the number of kernels for each length to 128. We also use dropout \cite{srivastava2014dropout} and L2 regularization to prevent over-fitting. Based on the obtained results, the official f1-score of CNN on \dataset~dataset is 69.28\%.

\subsubsection{Att-BLSTM}
We used one layer of bi-directional LSTM and set the hidden state size to 100. To prevent over-fitting, we use L2 regularization, recurrent dropout, and regular dropout. Based on the obtained results, the official f1-score of Att-BLSTM on \dataset~dataset is 69.61\%.
        
\subsubsection{BLSTM-LET}
We use four attention heads in the multi-head attention layer and set the layer size to 50 for each head. Hidden state of the LSTM is set to 300. Like the previous model, recurrent and regular dropout, as well as L2 regularization, are used. Based on the obtained results, the official f1-score of BLSTM with entity typing (BLSTM-LET) on \dataset~dataset is 70.79\%.

\subsubsection{R-BERT}
We fine-tune the base BERT pre-trained model for this method. Other hyper-parameters can be seen in Table~\ref{Hyper}. Based on the obtained results, the official f1-score of BLSTM with entity typing (BLSTM-LET) on \dataset~dataset is 75.31\%.

\subsubsection{BERTEM-MTB}
We fine-tune the base BERT pre-trained model for this method. Other hyper-parameters can be seen in Table~\ref{Hyper}. Based on the obtained results, the official f1-score of BLSTM with entity typing (BLSTM-LET) on \dataset~dataset is 77.66\%.

\begin{table*}
    \caption{Hyper-parameters for training R-BERT and BERTEM-MTB models}
    \label{Hyper}
     \begin{tabular}{@{\extracolsep{\fill}}lr}
	\hline
		Hyper-parameter & Value \\
    \hline
    Sentence length & 128                    \\ 
    Batch size      & 16                     \\ 
    Optimizer       & Adam                   \\ 
    Learning rate   & 3e-5                   \\ 
    \hline
    \end{tabular}
\end{table*}

\subsection{Results Per Classes}
The final results for individual classes can be seen in Table~\ref{CNN-each-class-tbl}. As can be seen, the f1 measure of the BERTEM-MTB model is higher than the other models in all classes. Also, in almost all classes, the value of f1 has risen from the lowest in Baseline to the highest in R-BERT. However, the models on the Instrument-Agency class do not behave the same, which means that the baseline model is better than all models except BERTEM-MTB. The reason for this is that the baseline model uses dependency relations and their direction between two consecutive entities for this purpose, while other models do not use this information. Sentences containing the Instrument-Agency relation class are very similar in terms of the dependency tree. Consequently, the baseline model, which uses dependency tree information, has learned how to detect this kind of relationship by observing a similar pattern.

\begin{table*}[ht!]

    \caption{f1-scores for the nine classes of \dataset~for all six models}
    \label{CNN-each-class-tbl}
	\begin{tabular}{@{\extracolsep{\fill}}lrrrrrr}
		\hline
		Class              & Baseline &   CNN   & Att-BLSTM & BLSTM-LET & R-BERT  &    BERTEM-MTB    \\ 
		\hline
		Cause-Effect       & 77.41\%  & 80.53\% &  82.13\%  &  81.06\%  & 83.51\% & \textbf{86.07\%} \\
		Component-Whole    & 53.22\%  & 62.56\% &  63.59\%  &  63.74\%  & 66.87\% & \textbf{67.09\%} \\
		Content-Container  & 70.31\%  & 75.84\% &  73.82\%  &  75.90\%  & 77.24\% & \textbf{78.53\%} \\
		Entity-Destination & 73.53\%  & 75.82\% &  79.24\%  &  81.05\%  & 84.77\% & \textbf{86.00\%} \\
		Entity-Origin      & 60.95\%  & 66.80\% &  66.80\%  &  66.92\%  & 74.95\% & \textbf{76.08\%} \\
		Instrument-Agency  & 64.81\%  & 57.78\% &  54.72\%  &  59.56\%  & 62.25\% & \textbf{69.51\%} \\
		Member-Collection  & 58.66\%  & 70.25\% &  67.56\%  &  69.55\%  & 75.17\% & \textbf{76.00\%} \\
		Message-Topic      & 59.07\%  & 71.94\% &  74.90\%  &  75.33\%  & 81.34\% & \textbf{85.19\%} \\
		Product-Producer   & 53.07\%  & 61.99\% &  63.74\%  &  64.02\%  & 71.67\% & \textbf{74.46\%} \\ 
		\hline
	\end{tabular}
\end{table*}

\section{Conclusion}
\label{conclusion}
In this paper, the Relation Extraxtion (RE) task in the Persian language is conducted for the first time. For this purpose, we initially proposed a bilingual version of the Semeval-2010-Task-8 dataset, dubbed as \dataset. Then, having investigated the state-of-the-art language-agnostic methods for RE in the English language, we adapted and customized some of these methods for the Persian language. Moreover, a logistic regression algorithm with syntactic and semantic features is employed as a baseline. The acquired experimental results not only double-proved the superiority of the BERT-based models over the baseline and other deep learning models but also proved its comparability to similar state-of-the-art methods in English. However, due to particular challenges in the processing of the Persian language such as being free-word-order and having ambiguity-prone nature in comparison with English, the performance of the customized methods in Persian was less than their performance in English.

As the future work of this study, more accurate Persian word embedding can be presented and applied to improve the results of non-BERT-based models. Moreover, by designing training steps tailored to the Persian language features, a novel BERT-based RE model can be proposed for the Persian language.

\section{Acknowledgments}
It is necessary to acknowledge the active collaboration of Dr. Sayyed Ali Hossayni and Mr. Kamyar Darvishi, who kindly collaborated with us, during the conduction of this research.

\bibliographystyle{unsrt}
\bibliography{perlex}

\begin{thebibliography}{10}

\bibitem{suchanek2007yago}
Fabian~M Suchanek, Gjergji Kasneci, and Gerhard Weikum.
\newblock Yago: a core of semantic knowledge.
\newblock In {\em Proceedings of the 16th international conference on World
  Wide Web}, pages 697--706. ACM, 2007.

\bibitem{bollacker2008freebase}
Kurt Bollacker, Colin Evans, Praveen Paritosh, Tim Sturge, and Jamie Taylor.
\newblock Freebase: A collaboratively created graph database for structuring
  human knowledge.
\newblock In {\em Proceedings of the 2008 ACM SIGMOD International Conference
  on Management of Data}, SIGMOD '08, pages 1247--1250. ACM, 2008.

\bibitem{auer2007dbpedia}
S\"{o}ren Auer, Christian Bizer, Georgi Kobilarov, Jens Lehmann, Richard
  Cyganiak, and Zachary Ives.
\newblock Dbpedia: A nucleus for a web of open data.
\newblock In {\em Proceedings of the 6th International The Semantic Web and 2Nd
  Asian Conference on Asian Semantic Web Conference}, ISWC'07/ASWC'07, pages
  722--735. Springer-Verlag, 2007.

\bibitem{vrandevcic2014wikidata}
Denny Vrande\v{c}i\'{c} and Markus Kr\"{o}tzsch.
\newblock Wikidata: A free collaborative knowledgebase.
\newblock {\em Communications of the ACM}, 57(10):78--85, September 2014.

\bibitem{asgari2019farsbase}
Majid Asgari-Bidhendi, Ali Hadian, and Behrouz Minaei-Bidgoli.
\newblock Farsbase: The persian knowledge graph.
\newblock {\em Semantic Web}, 10(6):1169--1196, 2019.

\bibitem{hendrickx2009semeval}
Iris Hendrickx, Su~Nam Kim, Zornitsa Kozareva, Preslav Nakov, Diarmuid
  {\'O}~S{\'e}aghdha, Sebastian Pad{\'o}, Marco Pennacchiotti, Lorenza Romano,
  and Stan Szpakowicz.
\newblock Semeval-2010 task 8: Multi-way classification of semantic relations
  between pairs of nominals.
\newblock In {\em Proceedings of the Workshop on Semantic Evaluations: Recent
  Achievements and Future Directions}, pages 94--99. Association for
  Computational Linguistics, 2009.

\bibitem{zhang2017position}
Yuhao Zhang, Victor Zhong, Danqi Chen, Gabor Angeli, and Christopher~D Manning.
\newblock Position-aware attention and supervised data improve slot filling.
\newblock In {\em Proceedings of the 2017 Conference on Empirical Methods in
  Natural Language Processing}, pages 35--45, 2017.

\bibitem{walker2006ace}
Christopher Walker, Stephanie Strassel, Julie Medero, and Kazuaki Maeda.
\newblock Ace 2005 multilingual training corpus.
\newblock {\em Linguistic Data Consortium, Philadelphia}, 57, 2006.

\bibitem{devlin2019bert}
Jacob Devlin, Ming-Wei Chang, Kenton Lee, and Kristina Toutanova.
\newblock Bert: Pre-training of deep bidirectional transformers for language
  understanding.
\newblock In {\em Proceedings of the 2019 Conference of the North American
  Chapter of the Association for Computational Linguistics: Human Language
  Technologies, Volume 1 (Long and Short Papers)}, pages 4171--4186, 2019.

\bibitem{lee2019semantic}
Joohong Lee, Sangwoo Seo, and Yong~Suk Choi.
\newblock Semantic relation classification via bidirectional lstm networks with
  entity-aware attention using latent entity typing.
\newblock {\em Symmetry}, 11(6):785, 2019.

\bibitem{mikolov2013efficient}
Tomas Mikolov, Kai Chen, Greg Corrado, and Jeffrey Dean.
\newblock {Efficient Estimation of Word Representations in Vector Space}.
\newblock In {\em Proceedings of the First International Conference on Learning
  Representations}, ICLR '13, pages 1--13, 2013.

\bibitem{soares2019matching}
Livio~Baldini Soares, Nicholas FitzGerald, Jeffrey Ling, and Tom Kwiatkowski.
\newblock Matching the blanks: Distributional similarity for relation learning.
\newblock In {\em Proceedings of the 57th Annual Meeting of the Association for
  Computational Linguistics}, pages 2895--2905, 2019.

\bibitem{han2018fewrel}
Xu~Han, Hao Zhu, Pengfei Yu, Ziyun Wang, Yuan Yao, Zhiyuan Liu, and Maosong
  Sun.
\newblock Fewrel: {A} large-scale supervised few-shot relation classification
  dataset with state-of-the-art evaluation.
\newblock In {\em Proceedings of the 2018 Conference on Empirical Methods in
  Natural Language Processing, Brussels, Belgium, October 31 - November 4,
  2018}, pages 4803--4809. Association for Computational Linguistics, 2018.

\bibitem{mintz2009distant}
Mike Mintz, Steven Bills, Rion Snow, and Dan Jurafsky.
\newblock Distant supervision for relation extraction without labeled data.
\newblock In {\em Proceedings of the Joint Conference of the 47th Annual
  Meeting of the ACL and the 4th International Joint Conference on Natural
  Language Processing of the AFNLP: Volume 2 - Volume 2}, ACL '09, pages
  1003--1011. Association for Computational Linguistics, 2009.

\bibitem{riedel2010modeling}
Sebastian Riedel, Limin Yao, and Andrew McCallum.
\newblock Modeling relations and their mentions without labeled text.
\newblock In {\em Joint European Conference on Machine Learning and Knowledge
  Discovery in Databases}, pages 148--163. Springer, 2010.

\bibitem{hoffmann2011knowledge}
Raphael Hoffmann, Congle Zhang, Xiao Ling, Luke Zettlemoyer, and Daniel~S.
  Weld.
\newblock Knowledge-based weak supervision for information extraction of
  overlapping relations.
\newblock In {\em Proceedings of the 49th Annual Meeting of the Association for
  Computational Linguistics: Human Language Technologies - Volume 1}, HLT '11,
  pages 541--550. Association for Computational Linguistics, 2011.

\bibitem{surdeanu2012multi}
Mihai Surdeanu, Julie Tibshirani, Ramesh Nallapati, and Christopher~D. Manning.
\newblock Multi-instance multi-label learning for relation extraction.
\newblock In {\em Proceedings of the 2012 Joint Conference on Empirical Methods
  in Natural Language Processing and Computational Natural Language Learning},
  EMNLP-CoNLL '12, pages 455--465. Association for Computational Linguistics,
  2012.

\bibitem{yang2019exploiting}
Kaijia Yang, Liang He, Xin-yu Dai, Shujian Huang, and Jiajun Chen.
\newblock Exploiting noisy data in distant supervision relation classification.
\newblock In {\em Proceedings of the 2019 Conference of the North {A}merican
  Chapter of the Association for Computational Linguistics: Human Language
  Technologies, Volume 1 (Long and Short Papers)}, pages 3216--3225.
  Association for Computational Linguistics, June 2019.

\bibitem{qin2018robust}
Pengda Qin, Weiran Xu, and William~Yang Wang.
\newblock Robust distant supervision relation extraction via deep reinforcement
  learning.
\newblock In {\em Proceedings of the 56th Annual Meeting of the Association for
  Computational Linguistics (Volume 1: Long Papers)}, pages 2137--2147.
  Association for Computational Linguistics, July 2018.

\bibitem{feng2018reinforcement}
Jun Feng, Minlie Huang, Li~Zhao, Yang Yang, and Xiaoyan Zhu.
\newblock Reinforcement learning for relation classification from noisy data.
\newblock In {\em Proceedings of the Thirty-Second {AAAI} Conference on
  Artificial Intelligence, (AAAI-18), the 30th innovative Applications of
  Artificial Intelligence (IAAI-18), and the 8th {AAAI} Symposium on
  Educational Advances in Artificial Intelligence (EAAI-18), New Orleans,
  Louisiana, USA, February 2-7, 2018}, pages 5779--5786. {AAAI} Press, 2018.

\bibitem{vashishth2018reside}
Shikhar Vashishth, Rishabh Joshi, Sai~Suman Prayaga, Chiranjib Bhattacharyya,
  and Partha Talukdar.
\newblock {RESIDE}: Improving distantly-supervised neural relation extraction
  using side information.
\newblock In {\em Proceedings of the 2018 Conference on Empirical Methods in
  Natural Language Processing}, pages 1257--1266. Association for Computational
  Linguistics, October-November 2018.

\bibitem{wang2018label}
Guanying Wang, Wen Zhang, Ruoxu Wang, Yalin Zhou, Xi~Chen, Wei Zhang, Hai Zhu,
  and Huajun Chen.
\newblock Label-free distant supervision for relation extraction via knowledge
  graph embedding.
\newblock In {\em Proceedings of the 2018 Conference on Empirical Methods in
  Natural Language Processing}, pages 2246--2255. Association for Computational
  Linguistics, October-November 2018.

\bibitem{lin2016neural}
Yankai Lin, Shiqi Shen, Zhiyuan Liu, Huanbo Luan, and Maosong Sun.
\newblock Neural relation extraction with selective attention over instances.
\newblock In {\em Proceedings of the 54th Annual Meeting of the Association for
  Computational Linguistics (Volume 1: Long Papers)}, volume~1, pages
  2124--2133. Association for Computational Linguistics, August 2016.

\bibitem{ye2019distant}
Zhi{-}Xiu Ye and Zhen{-}Hua Ling.
\newblock Distant supervision relation extraction with intra-bag and inter-bag
  attentions.
\newblock In {\em Proceedings of the 2019 Conference of the North American
  Chapter of the Association for Computational Linguistics: Human Language
  Technologies, {NAACL-HLT} 2019, Minneapolis, MN, USA, June 2-7, 2019, Volume
  1 (Long and Short Papers)}, pages 2810--2819. Association for Computational
  Linguistics, 2019.

\bibitem{rink2010utd}
Bryan Rink and Sanda Harabagiu.
\newblock {UTD}: Classifying semantic relations by combining lexical and
  semantic resources.
\newblock In {\em Proceedings of the 5th International Workshop on Semantic
  Evaluation}, pages 256--259. Association for Computational Linguistics, July
  2010.

\bibitem{yates2007textrunner}
Alexander Yates, Michael Cafarella, Michele Banko, Oren Etzioni, Matthew
  Broadhead, and Stephen Soderland.
\newblock Textrunner: open information extraction on the web.
\newblock In {\em Proceedings of Human Language Technologies: The Annual
  Conference of the North American Chapter of the Association for Computational
  Linguistics: Demonstrations}, pages 25--26. Association for Computational
  Linguistics, 2007.

\bibitem{renslow2018lightrel}
Tyler Renslow and G{\"u}nter Neumann.
\newblock Lightrel at semeval-2018 task 7: Lightweight and fast relation
  classification.
\newblock In {\em Proceedings of The 12th International Workshop on Semantic
  Evaluation}, pages 778--782, 2018.

\bibitem{zeng2014relation}
Daojian Zeng, Kang Liu, Siwei Lai, Guangyou Zhou, and Jun Zhao.
\newblock Relation classification via convolutional deep neural network.
\newblock In {\em Proceedings of {COLING} 2014, the 25th International
  Conference on Computational Linguistics: Technical Papers}, pages 2335--2344.
  Dublin City University and Association for Computational Linguistics, August
  2014.

\bibitem{zhou2016attention}
Peng Zhou, Wei Shi, Jun Tian, Zhenyu Qi, Bingchen Li, Hongwei Hao, and Bo~Xu.
\newblock Attention-based bidirectional long short-term memory networks for
  relation classification.
\newblock In {\em Proceedings of the 54th Annual Meeting of the Association for
  Computational Linguistics (Volume 2: Short Papers)}, volume~2, pages
  207--212. Association for Computational Linguistics, August 2016.

\bibitem{vaswani2017attention}
Ashish Vaswani, Noam Shazeer, Niki Parmar, Jakob Uszkoreit, Llion Jones,
  Aidan~N Gomez, \L~ukasz Kaiser, and Illia Polosukhin.
\newblock Attention is all you need.
\newblock In {\em Advances in Neural Information Processing Systems 30}, pages
  5998--6008. Curran Associates, Inc., 2017.

\bibitem{wu2019enriching}
Shanchan Wu and Yifan He.
\newblock Enriching pre-trained language model with entity information for
  relation classification.
\newblock In {\em Proceedings of the 28th ACM International Conference on
  Information and Knowledge Management}, pages 2361--2364. ACM, 2019.

\bibitem{poostchi2016personer}
Hanieh Poostchi, Ehsan~Zare Borzeshi, Mohammad Abdous, and Massimo Piccardi.
\newblock Personer: Persian named-entity recognition.
\newblock In {\em COLING 2016-26th International Conference on Computational
  Linguistics, Proceedings of COLING 2016: Technical Papers}, pages 3381--3389,
  2016.

\bibitem{srivastava2014dropout}
Nitish Srivastava, Geoffrey Hinton, Alex Krizhevsky, Ilya Sutskever, and Ruslan
  Salakhutdinov.
\newblock Dropout: a simple way to prevent neural networks from overfitting.
\newblock {\em The Journal of Machine Learning Research}, 15(1):1929--1958,
  2014.

\end{thebibliography}

\end{document}